\documentclass[runningheads]{llncs}
\usepackage[T1]{fontenc}
\usepackage{graphicx}
\usepackage{multirow}
\usepackage{booktabs}
\usepackage{subcaption}
\usepackage{cite}
\usepackage{amsmath}
\usepackage{verbatim}
\usepackage{float}
\usepackage{placeins}
\usepackage{hyperref}
\usepackage[toc,page]{appendix}

\newcommand\blfootnote[1]{%
  \begingroup
  \begin{NoHyper}
  \renewcommand\thefootnote{}\footnote{#1}%
  \addtocounter{footnote}{-1}%
  \endgroup
}

\begin{document}
\title{Informed Deep Abstaining Classifier:\\
Investigating noise-robust training for diagnostic decision support systems } 

\titlerunning{Informed Deep Abstaining Classifier}

\author{Helen Schneider\inst{1}*
\and Sebastian Nowak\inst{2}*
\and Aditya Parikh\inst{1}
\and Yannik C. Layer\inst{2}
\and Maike Theis\inst{2}
\and Wolfgang Block\inst{2}
\and Alois M. Sprinkart\inst{2} 
\and Ulrike Attenberger\inst{2}
\and Rafet Sifa\inst{1,3}}

%

\authorrunning{Schneider et al.}

\institute{Fraunhofer Institute for Intelligent Analysis and Information Systems IAIS\\ \email{ \{Helen.Schneider, Aditya.Parikh, Rafet.Sifa\} @iais.fraunhofer.de}  
\and
Department of Diagnostic and Interventional Radiology, University Hospital Bonn\\
\email{ \{Sebastian.Nowak, Alois\_.Sprinkart, Ulrike.Attenberger\} @ukbonn.de}
\and
Institute for Computer Science, University of Bonn\\
 *contributed equally
}


\maketitle              

\begin{abstract} 

Image-based diagnostic decision support systems (DDSS) utilizing deep learning have the potential to optimize clinical workflows. However, developing DDSS requires extensive datasets with expert annotations and is therefore costly. Leveraging report contents from radiological data bases with Natural Language Processing to annotate the corresponding image data promises to replace labor-intensive manual annotation. As mining 'real world' databases can introduce label noise, noise-robust training losses are of great interest. However,  current noise-robust losses do not consider noise estimations that can for example be derived based on the performance of the automatic label generator used. In this study, we expand the noise-robust Deep Abstaining Classifier (DAC) loss to an Informed Deep Abstaining Classifier (IDAC) loss by incorporating noise level estimations during training. Our findings demonstrate that IDAC enhances the noise robustness compared to DAC and several state-of-the-art loss functions. The results are obtained on various simulated noise levels using a public chest X-ray data set. These findings are reproduced on an in-house noisy data set, where labels were extracted from the clinical systems of the University Hospital Bonn by a text-based transformer. 
The IDAC can therefore be a valuable tool for researchers, companies or clinics aiming to develop accurate and reliable DDSS from routine clinical data.

\end{abstract}

\blfootnote{This research has been funded by the Federal Ministry of Education and Research of Germany and the state of North-Rhine Westphalia as part of the Lamarr-Institute for Machine Learning and Artificial Intelligence, LAMARR22B}

\section{Introduction}
Implementing artificial intelligence in medical workflows can enhance reading accuracy and productivity, thereby improving the quality and economic efficiency of patient care \cite{knapic, mango}. Effective and reliable DDSS, that aid physicians in detecting and characterizing pathologies, are therefore of great interest \cite{hosny}. Image-based deep learning models, e.g. Convolutional Neural Networks (CNNs), have shown high potential for detecting pathological alterations in medical imaging achieving human-level performance or exceeding humans \cite{serte, nowakliver1, nowakliver2, schneider2023symmetry, Maike_myom}. To develop a model with sufficient performance using supervised learning, a significant number of annotated training images are required. This is a major challenge in the medical field as the annotation of medical images requires considerable expertise from medical annotators and is therefore costly.\\

In in-house clinical data systems of radiology departments, diagnoses and findings are already documented by experts in free-text reports during the patients treatment. This information may be utilized as annotations of the associated medical images, eliminating or a least reducing the need for expensive manual annotation \cite{nowak2}. Consequently, analysing local databases of clinical routines is a promising way to develop models with sufficient data for reliable DDSS that offer real benefits in daily clinical practice. In recent years, this has been the motivation for various studies investigating the use of Natural Language Processing (NLP) to extract free-text report content as image labels. Further information about this specific NLP application and the challenges of medical semantics can be found in \cite{thirunavukarasu_large_2023, irvin, nowak1, nowak2, jain}.    \\

However, such labels can feature discrepancies to the image content due to various reasons, e.g. due to the error rate of the used NLP approach \cite{nowak2, jain}. In the following, labels with imperfections are referred to as noisy labels. It is common practice that the performance of labeling systems, which are either created for a medical-scientific project or implemented in a clinical workflows, are evaluated on independent and manually curated test data \cite{irvin, nowak1, nowak2}. This evaluation can be used to estimate the noise of the labels generated by the annotation system. A significant proportion of noisy labels can greatly impact the model's performance, posing potential harm to patients when used as DDSS. \\

Noise-robust algorithms, which can achieve exceptional performance despite label noise, represent therefore a highly relevant research field in medical image processing. Due to extensive data-protected image data sets and limited access to comprehensive local computing capacity, a desirable algorithm feature is a trivial implementation with no increased training complexity. Among other things, increased numbers of hyper- and network parameters can raise the training complexity. The expected label noise, given in practice due to the performance evaluation of the automatic labeler, can present valuable prior knowledge for these solution approaches.\\

Due to these requirements, noise-robust and easy-to-apply algorithms are of great interest to researchers, companies, and clinics aiming to develop DDSS with routine clinical data. 

\subsection{Related Work}
When dealing with data sets containing noisy labels, there are generally two non-mutually exclusive approaches: noise-robust training and noise detection. In noise-robust training modules are added that enable effective training of the network even in the presence of label noise. This involves implementing modules in network architectures \cite{robust_architecture_1, robust_architecture_2, robust_architecure_3}, developing co-teaching approaches \cite{han2018co} or applying resilient regularization \cite{regularization_1, regularization_2}, among other strategies. Additionally, noise-robust loss functions \cite{NEURIPS2023_activenegative, robust_loss_function_1, robust_loss_function_2, schneider2023one, robust_loss_function_4, symmetric_loss, asymmetricloss} represent established and easy-to-apply solution approaches.\\ 

Noise detection methods aim to identify noisy samples which enables excluding noisy samples in a new training iteration \cite{thulasidasan}, treating them as unlabeled data \cite{wang2022scalable} or efficient targeted relabeling \cite{niehues}. Noisy samples are detected by conspicuous patterns such as high model and label disagreement \cite{niehues}, gradient direction \cite{ren2018learning}, or disagreement between different models. Note that these approaches require a two-step training scheme, and are therefore increasing the training effort. \\

Thulasidasan et al. introduced the Deep Abstain Classifier (DAC) that can abstain from noisy labels in the loss function by enabling the network to map potential noisy samples to an extra neuron of the classification head \cite{thulasidasan}. High performances were achieved without complex hyperparameter searches and increased numbers of network parameters. 
However, the abstain mechanism does not incorporate prior knowledge of expected noise levels in the dataset, which could be valuable information for enhancing its performance. Furthermore, most methods are not sufficiently tested with realistic noise simulations and real-world noisy data sets for medical imaging.\\\\

It is, therefore, the aim of this work:
\begin{itemize}
\item to propose an extension of the DAC, called \textbf{Informed Deep Abstaining Classifier (IDAC)}, that enables the inclusion of prior knowledge on the expected noise inside the abstaining loss function. The IDAC represents an easy-to-apply noise-robust loss, leading to no significant increase in the training complexity. 
\item to evaluate the noise-robust training of the algorithm compared to state-of-the-art baseline loss functions on a public medical dataset with realistic simulated noise levels. We focus on low noise levels between $1$\%-$15$\%, which have not been sufficiently researched in the existing literature despite their high practical relevance.
\item to also evaluate IDAC on a real-world in-house data set of clinical routine with automatically generated labels, underlining the impact of the IDAC for the development of DDSS system with clinical routine data.
\end{itemize}

\section{Materials and Methods}

\subsection{Data Sets}

\subsubsection{Public data set} 
The experiments are based on CheXpert, a chest X-ray data set with automatically generated training labels based on the corresponding text reports \cite{irvin}. We create two training data sets for binary classification (diseased vs. non-diseased) for chest X-ray images in posterior-anterior projection. We choose pleural effusion (n=$37530$) and cardiomegaly (n=$7637$) to investigate manifestations of pulmonary and cardiovascular diseases. Both conditions are well-represented in the CheXpert data set, with low intrinsic noise due to high-quality labeling by the NLP label system \cite{irvin}. We exclude the uncertainty labels of the training data to further minimize the intrinsic noise. Various noise levels are simulated by swapping correct with incorrect labels \mbox{(1\%, 3\%, 5\%, 7\%, 15\%, 30\%, 50\%)}. The validation set contains $202$ and the testing set contains $500$ manually annotated samples.

\subsubsection{Clinical Real-World data set}
We also explore the proposed approach using an in-house data set of chest X-ray images of the intensive care units of the hospital Bonn in anterior-posterior projection annotated by report content extracted in previous studies \cite{nowak1, nowak2}. Here, only pleural effusion is investigated, as no cardiomegaly labels are available. The labels of the training set (n=$10000$) are based on report content extracted by a BERT transformer \cite{Devlin2019BERTPO}. Further information about utilized NLP methods are available in \cite{nowak1}. The labels of the validation set (n=$1437$) are based on report content extracted by research assistants. The test set (n=$187$) is labeled based on the image data by a medical expert. Noise levels were estimated in previous studies. There is a $4.3\%$ discrepancy in a subset between the labels based on report content compared to labels based on the images themselves and a $1.2\%$ discrepancy caused by errors by the BERT transformer. Further details on the data set can be found in the previous open-access works \cite{nowak1, nowak2}. 

\subsection{Loss Functions}
In this section all relevant loss functions and corresponding classifiers for training k-class classification problems are given. Here $p_w(y=i|x)$ is the probability of the i-th class of a model output $y$ for a given input image $x$ dependent on the model weights $w$. For simplicity $p_i$ is used instead of $p_w(y=i|x)$ and $t_i$ stands for the corresponding label, as in a previous work \cite{thulasidasan}. When representing the loss function for an entire batch, $p_{j,i}$ corresponds to the probability of the i-th class in the model output for the j-th sample within the considered batch.

\subsubsection{Cross Entropy Loss}
The standard cross entropy (CE) loss is defined as followed,
\begin{equation}
\mathcal{L}_{\text{CE}}(x_j) = -\sum_{i=1}^k t_i \log (p_i)
\end{equation}
where  $t_i$ is ground truth of the sample $x_j$ for class $i \leq k$ . Although the CE loss performs exceptionally well in clean label training scenarios, it lacks noise robustness, as noted in \cite{symmetric_loss}. 

\subsubsection{Symmetric Cross Entropy Loss}
The symmetric cross entropy (SCE) loss was introduced to improve the noise robustness of the CE loss without structural adaption of the classifier itself \cite{symmetric_loss}. The SCE adds a Reverse Cross Entropy (RCE) term with clipping of predictions and true targets \cite{symmetric_loss}. 
It is defined as followed:
$$\mathcal{L}_{\text{SCE}}(x_j) = -\sum_{i=1}^k t_i \log (p_i) - \sum_{i=1}^k p_i \log (t_i) = \mathcal{L}_{\text{CE}}(x_j) + \mathcal{L}_{\text{RCE}}(x_j)$$

\subsubsection{Deep Abstaining Classifier Loss}
The DAC has an additional $k+1$ output neuron $p_{k+1}$, designed to represent the probability of abstention for the data point $x_j$ during training \cite{thulasidasan}. The DAC loss is defined as followed:
$$\mathcal{L}_{\text{DAC}}(x_j) = (1 - p_{k+1}) \left( -\sum_{i=1}^k t_i \log \frac{p_i}{1-p_{k+1}} \right) + \alpha \log \frac{1}{1-p_{k+1}}$$
The first term defines an adjustment of the CE loss over the $k$ non-abstaining classes. If the abstaining output is absent ($p_{k+1}=0$), it reverts to the standard CE loss, otherwise, the $k$ class probabilities are normalized based on the output of the abstain neuron $p_{k+1}$. The second term with abstention weight $\alpha>0$ is a regularization term to penalize abstention, avoiding the DAC to abstain from all training cases. \\
$\alpha$ is linearly increased during training based on the maximal number of training epochs. This allows fewer and fewer cases to be abstained during training, initially focusing on the supposedly clean labels. However, a complex dependency between the loss function and the maximal number of training epochs is introduced, leading to costly hyperparameter searches and less intuitive training analysis.  \\

\subsubsection{Informed Deep Abstaining Classifier Loss}
The IDAC loss represents an enhancement of the DAC loss function by taking into account information about the expected label noise of the data set. The classifier, like the DAC, has an additional neuron $p_{k+1}$, which represents the probability of abstention for a sample $x_j$.
While the true label noise percentage $\eta$ is not known in practice, often an estimate of the noise is given. In the following the noise estimation is referred to as $\Tilde{\eta}$.
In practical implementation, $\Tilde{\eta}$ can be based on the performance evaluation of the annotators (both for automatic NLP and human annotators), which are typically performed on subsets of the medical dataset. The noise estimation $\Tilde{\eta}$ represents a fixed hyperparameter for the loss function and is not adapted during training. \\ The IDAC loss extension aims to take this prior knowledge of the data set into account during the data-driven training. The IDAC and DAC loss are methodically differentiated by the abstain regularization term. The IDAC loss is  therefore defined as followed:

\begin{align*}
    \mathcal{L}_{\text{IDAC}}(x_j) &=  (1 - p_{k+1}) \left( -\sum_{i=1}^k t_i \log \frac{p_i}{1-p_{k+1}} \right) + \alpha(\Tilde{\eta}-\hat{\eta})^2 \\
    \hat{\eta} &=\sum_{l=1}^N \frac{p_{l,k+1}}{N}
\end{align*}

where $N$ represents the batch size and $\hat{\eta}$ is the currently applied abstention of the classifier per batch. As the direct computation of the ratio of abstained data points in a batch is not differentiable (due to argmax), $\hat{\eta}$ is approximated by summing the softmax outputs of the abstaining neuron with division by $N$. Note that the noise estimation $\Tilde{\eta}$ is included in the regularization term. \\
If the model abstains to many samples during training compared to the noise estimation $\Tilde{\eta}$, the regularization term increases and leads to a stronger penalization for abstaining. This can result in a reduction of abstained samples, ideally excluding correct (but difficult to learn) training samples during the abstention.
A too-low abstention rate compared to the noise estimation $\Tilde{\eta}$ during training also results in a high regularization term. Minimizing the penalty term leads to an increase of abstained samples. The goal is to exclude more noisy label samples during training to minimize the effects of noise overfitting on the training performance.\\
Compared to the DAC loss, $\alpha$ remains constant and is independent of the number of training epochs. This can lead to a more user-friendly training process and a less complex hyperparameter search. Additionally, the utilized model does not have a significant increase of parameters, as only a single additional output neuron is required. The IDAC therefore represents a easy-to-implement and easy-to-apply noise-robust classifier, that does not require a more complex training schedule (e.g. a significantly increased number of training weights, hyperparameters, or training runs).\\
Note that due to the inclusion of the expected percentage of noise per batch in the IDAC loss, we recommend implementing high batch sizes during IDAC training to obtain a more realistic representation of the noise distribution per batch.

\subsection{Experiments}

We apply DenseNet-121 \cite{densnet} with ImageNet \cite{imagenet} pretrained weights from PyTorch as an established model for processing lung diseases in chest X-ray images \cite{irvin, schneider2022esann}. Motivated by the DAC paper\cite{thulasidasan}, the model is trained with the stochastic gradient descent for $300$ epochs with a momentum of $0.9$, a weight decay of $5e{-4}$ and a batch size of 512. The learning rate of $0.1$ is decreased at epoch 100 and 250 by factor of $0.1$. For the pleural effusion subset of CheXpert $500$ epochs are trained to ensure convergence. All images are ImageNet normalized, reshaped to the size $224 \times 224$ and augmented by affine transformations during training. As CNNs tend to learn relevant features during the initial training phase, even with noisy labels \cite{zhu2021understanding}, the IDAC and DAC are first trained by applying the CE loss to all $k+1$ output neurons (including the abstention neuron) in a warm-up phase. For classifying pleural effusion in the CheXpert data set the warm-up phase is set to 10 epochs. As the in-house pleural effusion and the CheXpert cardiomegaly data sets feature less training samples, we perform a hyperparameter search for DAC and IDAC ($\text{warm-up} \in \{10,30,50\} \text{ epochs}$). Additionally, the hyperparameter ($\alpha \in \{ 1, 10, 20\}$) is included in the search to investigate the influence of $\alpha$ when training IDAC with different noise levels. \\

For all simulated noise experiments, the IDAC loss utilizes the percentage of simulated noise as $\Tilde{\eta}$. Note that this represents only a noise estimation due to the inherent noise of the CheXpert data set. For the analysis of $0$\% simulated noise, we use an estimated noise level $\Tilde{\eta}$ of $0.5$\% due to the inherited noise of the CheXpert dataset.
For training with the clinical in-house data set, we employ an upper estimate of $5\%$ for the expected noise.\\
For both IDAC and DAC the abstain neuron $p_{k+1}$ is excluded from inference on the validation and test set, as we assume that these manually annotated sets do not include noisy labels. \\

To enable a fair comparison, we focus on noise-robust loss functions as baseline functions. Other state-of-the-art methods that lead to an increased number of training parameters \cite{han2018co}, several additional hyperparameters \cite{wang2024tackling}, or are an aggregation of multiple techniques \cite{nguyen2019self, li2020dividemix} do not, in our opinion, provide a fair comparison to the easily tuned and trained IDAC loss. We therefore implement the CE and SCE loss, as well as the DAC as baseline. 
For the less extensive cardiomegaly data set we additionally analyzed the performance of the normalized cross entropy (NCE), the normalized generalized cross entropy (NGCE) and asymmetric generalized cross entropy (AGCE) loss. Although they achieved exceptional performance on commonly used computer vision baseline datasets like CIFAR10 \cite{asymmetricloss}, we were unable to achieve comparable performance to the other baseline functions for the more complex medical image use case after a reasonable grid search. The obtained results and implemented hyperparameter searches are included in the appendix \ref{appendix}.\\

For each investigated loss, models are selected based on the validation area under the receiver operating characteristics curve (AUROC) and compared by test AUROC. We calculate the 95\% confidence intervals (CI) using bootstrapping with 1000 resamples to improve the performance evaluation. Training and evaluation are conducted on a NVIDIA A100 40GB GPU with CUDA Toolkit 11.3, PyTorch v1.12.1, TorchMetrics v0.10.3 and Scikit-learn v0.24.2.

\section{Results}

\subsubsection{Noise Robustness}
\begin{figure}[!h]
        \centering
        \includegraphics[width=0.6\textwidth]{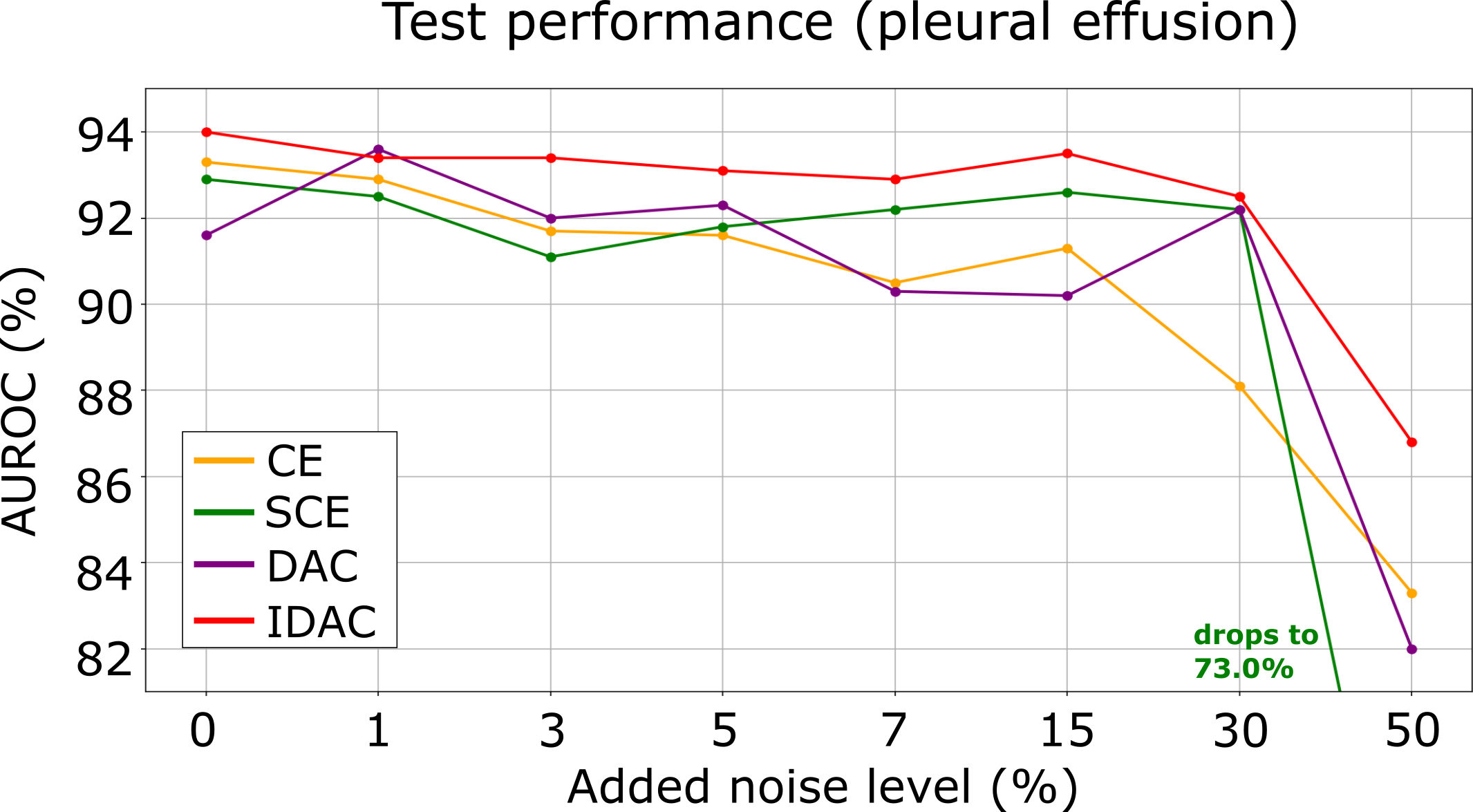}
        \caption{
        AUROC scores (\%) on the CheXpert test set for simulated noise levels below 50\% for the detection of pleural effusion. The proposed IDAC loss achieves the highest noise robustness. Note that we only visualize scores higher than 80\% due to overview reasons.}
        \label{fig:noise_vs_performance_1}
\end{figure}

AUROC scores on the hold-out test sets of the \mbox{CheXpert} data are given for all investigated losses trained on different simulated noise levels for all use cases in Table \ref{tab:evaluation_table_pf}, Table \ref{tab:evaluation_table_cardio}, Figure \ref{fig:noise_vs_performance_1} and Figure \ref{fig:noise_vs_performance_2}. \\

As expected the performance decreases in general for all loss function with increasing noise levels. This is most effectively mitigated by the proposed IDAC loss. Here, IDAC training achieves the highest performance for classifying pleural effusion and cardiomegaly in six out of seven noise levels.\\ 

For the two cases where IDAC does not reach the highest score, it still outperforms two of the three baselines, there is only a small discrepancy to the best score. This indicates that the IDAC training has the potential to improve noise robustness, for data sets of both substantial and smaller sizes. The direct comparison to the DAC performance underlines that the abstaining loss functions benefit from the prior knowledge of the expected noise.\\
It can be seen, that low noise levels like $3$\% reduce the CE performance, stressing the importance of analyzing practically relevant low noise levels below 15\% of noise for noise-robust training in medical image analysis. If there is no simulated noise, a very low inherited noise level can be expected. The IDAC loss outperforms the considered baselines by up to $2.4$\%, stressing its usability to work with low noise levels.\\
For the high noise level $50$\% a significant improvement can be achieved by the cardiomegaly IDAC approach compared to the commonly used CE and SCE loss functions. \\

\begin{table}[!h]
    \centering
    \begin{tabular}{c@{\hspace{6pt}}|@{\hspace{6pt}}c@{\hspace{12pt}}c@{\hspace{12pt}}c@{\hspace{12pt}}c@{\hspace{6pt}}}
    \toprule
    Noise (\%) & CE                & SCE               & DAC                        & IDAC                       \\
    \midrule
    0     & 93.3 [91.2, 95.6] & 92.9 [90.3, 95.0] & 91.6 [88.8, 94.1]          & \textbf{94.0} [92.0, 95.7] \\
    1     & 92.9 [90.5, 95.1] & 92.5 [90.0, 95.1] & \textbf{93.6} [90.9, 95.7] & 93.4 [91.2, 95.4]          \\
    3     & 91.7 [89.3, 94.1] & 91.1 [87.9, 93.6] & 92.0 [88.7, 94.3]          & \textbf{93.4} [91.1, 95.0] \\
    5     & 91.6 [89.2, 93.6] & 91.8 [89.8, 93.6] & 92.3 [89.4, 94.6]          & \textbf{93.1} [91.0, 95.5] \\
    7     & 90.5 [87.4, 93.2] & 92.2 [89.6, 95.0] & 90.3 [87.8, 92.7]          & \textbf{92.9} [90.6, 95.4] \\
    15    & 91.3 [87.3, 94.8] & 92.6 [89.7, 95.0] & 90.2 [87.0, 92.9]          & \textbf{93.5} [91.7, 95.3] \\
    30    & 88.1 [84.9, 91.4] & 92.2 [89.4, 95.0] & 92.2 [90.0, 94.8]          & \textbf{92.5} [90.6, 94.8] \\
    50    & 83.3 [79.5, 86.3] & 73.0 [68.2, 78.8] & 82.0 [78.5, 85.0]          & \textbf{86.8} [83.3, 90.2] \\
    \bottomrule
    \end{tabular} \vspace{1em}
    \caption{AUROC scores (\%) and confidence intervalls on the CheXpert test set for different simulated noise levels (\%) for pleural effusion. The highest AUROC scores are highlighted in bold. The proposed IDAC loss achieves the highest noise robustness.}
    \label{tab:evaluation_table_pf}
\end{table}

\begin{table}[!h]
    \centering
    \begin{tabular}{c@{\hspace{6pt}}|@{\hspace{6pt}}c@{\hspace{12pt}}c@{\hspace{12pt}}c@{\hspace{12pt}}c@{\hspace{6pt}}}
    \toprule
    Noise (\%) & CE                & SCE               & DAC                        & IDAC                       \\
    \midrule
    0    & 83.4 [79.6, 87.0] & 82.4 [78.7, 85.5]          & 80.1 [75.8, 83.5] & \textbf{84.1} [80.5, 88.0] \\
    1    & 79.2 [73.5, 83.9] & 79.9 [75.0, 84.9]          & 79.5 [75.5, 83.2] & \textbf{86.5} [83.1, 89.8] \\
    3     & 78.4 [74.3, 84.1] & 80.4 [77.1, 83.9]          & 81.7 [78.0, 85.1] & \textbf{82.6} [79.8, 86.4] \\
    5     & 81.1 [76.2, 85.2] & 79.1 [74.3, 83.6]          & 84.0 [80.7, 87.4] & \textbf{85.0} [81.7, 87.8] \\
    7     & 77.2 [72.3, 81.3] & 79.8 [76.1, 83.9]          & 79.3 [76.1, 83.0] & \textbf{82.2} [78.4, 85.9] \\
    15    & 73.7 [69.3, 77.5] & \textbf{78.6} [75.3, 82.4] & 74.0 [70.6, 78.0] & 78.1 [74.5, 82.6]          \\
    30    & 67.0 [62.3, 72.6] & 63.4 [58.2, 68.6]          & 68.7 [62.9, 73.1] & \textbf{74.4} [70.3, 79.3] \\
    50    & 59.1 [53.9, 63.2] & 66.6 [61.4, 71.8]          & 67.8 [63.9, 72.2] & \textbf{72.6} [68.1, 76.9] \\
    \bottomrule
    \end{tabular} \vspace{1em}
    \caption{AUROC scores (\%) and confidence intervalls on the CheXpert test set for different simulated noise levels (\%) for cardiomegaly. The highest AUROC scores are highlighted in bold. The proposed IDAC loss achieves the highest noise robustness.}
    \label{tab:evaluation_table_cardio}
\end{table}

\begin{figure}[!h]
        \centering
        \includegraphics[width=0.6\textwidth]{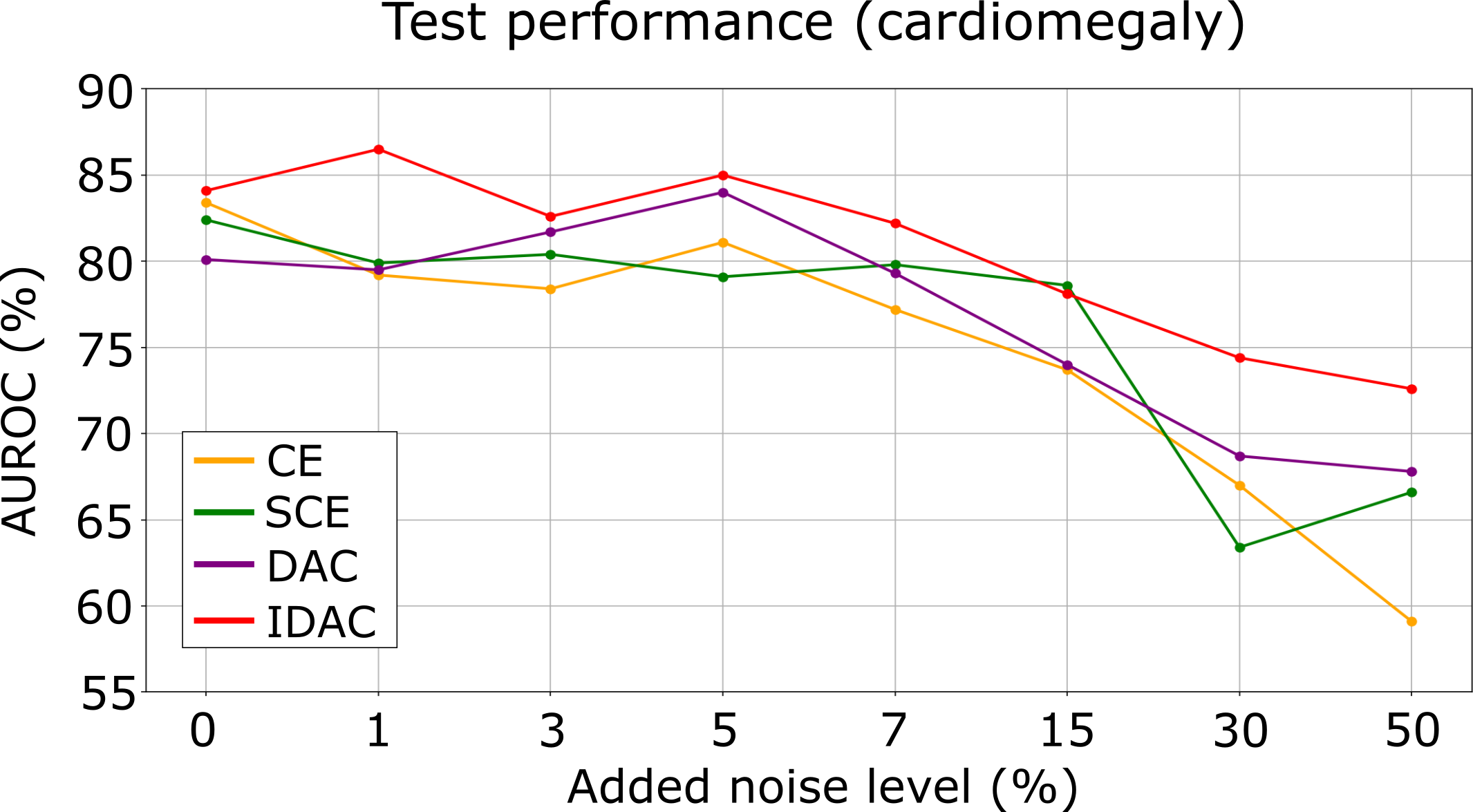}
        \label{fig:noise_vs_performance_2}
    
    \caption{
        AUROC scores (\%) on the CheXpert test set for simulated noise levels below 50\% for the detection of cardiomegaly. The proposed IDAC loss achieves the highest noise robustness. }
        \label{fig:noise_vs_performance_2}
\end{figure}

Table \ref{tab:in-house_performance} shows detailed performance metrics for all investigated losses on the in-house data set of clinical routine. IDAC training achieves the highest performance for classifying pleural effusion in four out of five performance metrics. Compared to conventional CE training, the AUROC score can be improved by up to 5\%. The results underline the strong noise-robustness of the IDAC loss, even for real-life noisy data sets. The IDAC can contribute to more reliable DDSS with datasets from clinical routine without additional implementation effort. The inclusion of further clinical in-house data experiments with various noise levels represent future work, to strengthen the evaluation of the IDAC loss.\\

\begin{table}[!h]
    \centering
    \begin{tabular}{
    c@{\hspace{6pt}}
    c@{\hspace{6pt}}
    c@{\hspace{6pt}}
    c@{\hspace{6pt}}
    c@{\hspace{6pt}}
    c
    }
        \toprule
        \multicolumn{6}{c}{Pleural Effusion} \\
        \midrule
        Loss & AUROC & Bal. Acc. & F1 score & Precision & Recall \\ 
        \midrule
        CE & 78.7 & 71.5 & 77.3 & 72.4 & 82.9 \\ 
        SCE & 80.8 & 71.1 & 78.0 & 73.6 & 82.9 \\ 
        DAC & 80.4 & 72.4 & 79.8 & 72.3 & \textbf{89.2} \\ 
        IDAC & \textbf{83.8} & \textbf{76.4} & \textbf{81.7} & \textbf{76.0} & 88.3 \\
        \bottomrule
    \end{tabular}
    \vspace{1em}
    \caption{Performance metrics (\%) on the test sets of the noisy in-house data set from clinical routine. The highest scores, overall achieved by the proposed IDAC loss, are highlighted in bold.}
    \label{tab:in-house_performance}
\end{table}

\subsubsection{Influence of the abstention weight $\alpha$:}
\begin{figure}[H]
\includegraphics[width=\textwidth]{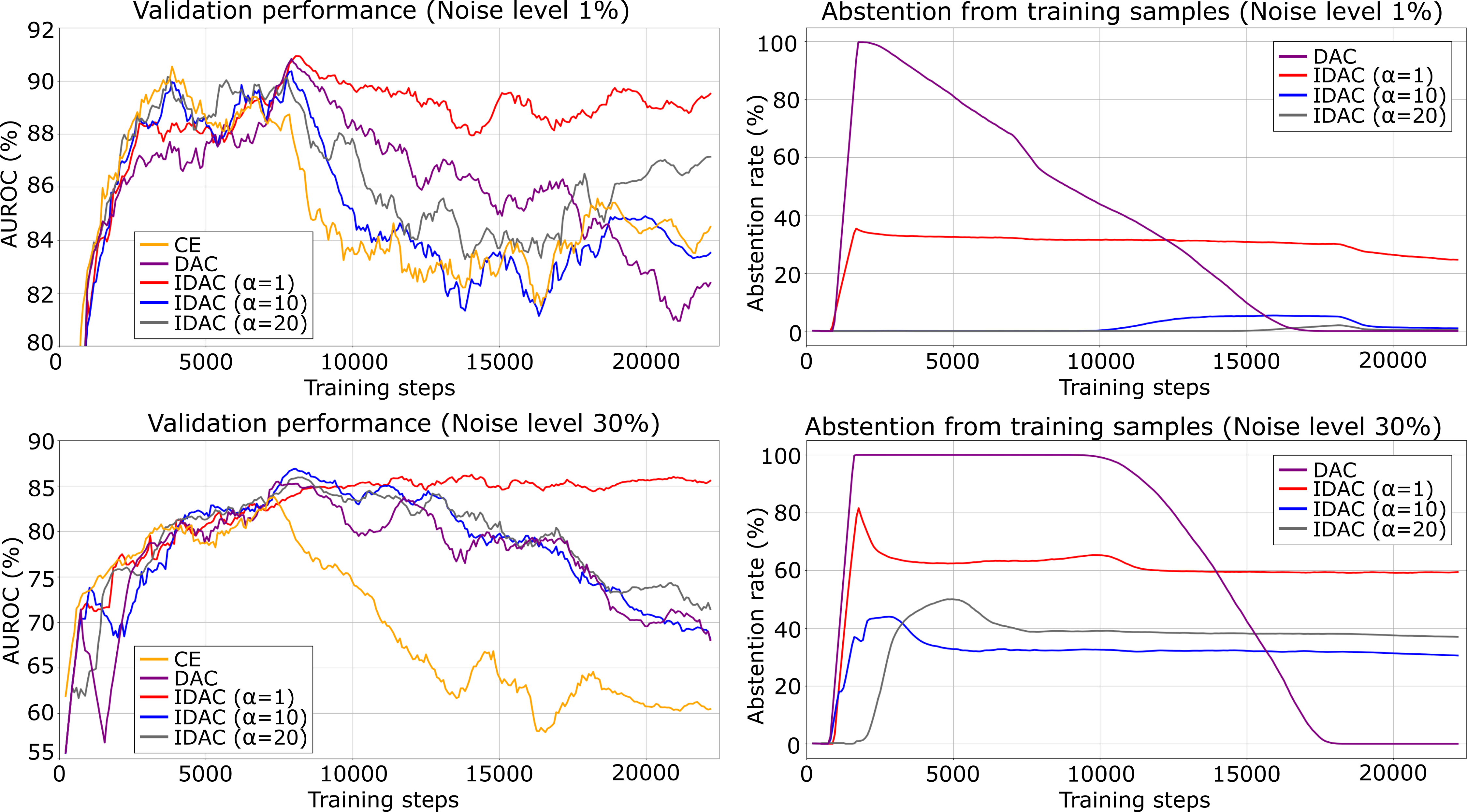}
\caption{Left: Smoothed validation performance of proposed IDAC training for classifying pleural effusion for CheXpert. Different abstention weights $\alpha$ for estimated noise levels $\Tilde{\eta}=1\%$ (upper) and $\Tilde{\eta}=30\%$ (lower) are investigated. Warm-up without abstention is 740 steps. For comparison, training with CE is illustrated in yellow and with DAC in purple. Right: Smoothed abstention rate of IDAC and DAC training with the different weight parameters $\alpha$ for the noise levels $1\%$ (upper) and $30\%$ (lower). The loss abstains from a training case when the model outputs the highest probability on the  $p_{k+1}$ abstention neuron after softmax.} \label{fig:2}
\end{figure}

Figure \ref{fig:2} shows abstention rates of the IDAC training for classifying pleural effusion on the \mbox{CheXpert} data set. Initially, more training cases are abstained than specified by the noise estimate $\Tilde{\eta}$. In the course of the training run, the abstention rate decreases and approaches $\Tilde{\eta}$. It can be observed that with a lower abstention regularization ($\alpha=1$) higher abstention rates occur during training, which results in a lower tendency to overfit compared to IDAC training with higher $\alpha$ or conventional CE training. The reduced tendency to overfit leads to an easier application and therefore higher usability of IDAC. However, for the high noise levels $(\Tilde{\eta}=30\%)$ slight performance benefits prior to overfitting are observed when training with high $\alpha=10$. As $(\Tilde{\eta}=30\%)$ allows for excessive abstention rates at the beginning of the training, a high regularization seems beneficial. These results indicate that the IDAC should be applied with low $\alpha$ regularization for lower noise levels and with higher $\alpha$ for high expected label noise.\\
The IDAC abstention behavior contrasts with the DAC training. Initially, above 90\% training data is abstained for the DAC loss. As training progresses and the regularization weight increases, this proportion decreases until eventually all samples are considered during training. Therefore the DAC training leads to lower classification performance and overfitting of the training data is observed. This suggests an advantage of integrating noise estimations into noise-robust training, particularly within abstaining classifiers. The results underline the superiority of the proposed IDAC loss.

\section{Conclusion}
Noise-robust training is key for the development of deep learning based DDSS from routine clinical image data. This work therefore introduces the noise-robust IDAC loss that incorporates an estimation of the expected noise during training. The IDAC improves noise robustness when evaluated for both simulated and real-world noisy medical imaging datasets.
Compared to conventional CE training, the AUROC score improves by 5\% for experiments on in-house clinical data and up to 7\% for simulated noise scenarios. Besides higher performance, the IDAC offers a simple implementation without increasing model complexity and greater usability by lower tendency to overfit. The IDAC can therefore be a valuable tool for researchers, companies or clinics that want to develop accurate and reliable DDSS from routine clinical data. 

The further development of the IDAC from multi-class classification to the medically highly relevant multi-label classification represents future work. Additionally, we want to investigate the influence of inaccurate noise estimations to further evaluate the robustness of the introduced approach.

\section*{Appendix}\label{appendix}
\subsection*{Additional Baseline Analysis}
We analyze the potential baselines normalized cross entropy (NCE)\cite{ma2020normalized}, normalized generalized cross entropy (NGCE) \cite{ma2020normalized} and asymmetric generalized cross entropy (AGCE) loss \cite{asymmetricloss} for the cardiomegaly data set and a simulated noise level of $15$\%. For each loss we implemented a reasonable grid search with the following hyperparameters:
\begin{itemize}
    \item NC: learning rate $ \in [0.1, 0.01]$
    \item NGCE:  learning rate $\in [0.1, 0.01]$, $q \in [0.25, 0.5, 0.75]$
    \item AGCE: learning rate $\in [0.1, 0.01]$, $q \in [0.25, 0.75, 1.25]$, $a \in [0.25, 0.75, 1.25]$
\end{itemize}
The implementations are based on the code available in \cite{asymmetricloss}.
Due to the low performance for the more complex medical image use cases, the loss functions NC, NGCE, and AGCE are not considered as baselines going forward.

\begin{table}[!h]
    \centering
    \begin{tabular}{c@{\hspace{6pt}}|@{\hspace{6pt}}c@{\hspace{12pt}}c@{\hspace{12pt}}c@{\hspace{12pt}}c@{\hspace{12pt}}|c@{\hspace{12pt}}c@{\hspace{12pt}}c@{\hspace{6pt}}}
    \toprule
    Noise (\%) & CE & SCE & DAC & IDAC & NC & NGCE & AGCE \\
    \midrule
    15   & 73.7 & \textbf{78.6} & 74.0 & 78.1 & 48.8 & 53.5 & 55.3 \\
    \bottomrule
    \end{tabular}
    \vspace{1em}
    \caption{ AUROC scores (\%) on the CheXpert test set for a simulated noise level of 15\% for the detection of cardiomegaly. The loss functions NCE, NGCE, and AGCE achieve comparatively low performance scores.}
    \label{tab:appendix}
\end{table}
\newpage
\bibliographystyle{splncs04}
\bibliography{bibliography}

\end{document}